\documentclass[10pt]{article}

\usepackage{amsmath,amsfonts} 

\def\eqref#1{equation~\ref{#1}}

\def\1{\bm{1}}

\DeclareMathAlphabet{\mathsfit}{\encodingdefault}{\sfdefault}{m}{sl}
\SetMathAlphabet{\mathsfit}{bold}{\encodingdefault}{\sfdefault}{bx}{n}

\usepackage{arxiv_style}
\usepackage{graphicx}
\usepackage{tabularx}
\usepackage{booktabs}
\usepackage{subcaption}
\usepackage{hyperref,url,array,xcolor,flafter}
\newcommand{\minite}{\textsc{miniTE}}
\newcommand{\learner}{\pi^{L}_{\theta}}
\newcommand{\rollout}{\pi^{R}_{\theta}}
\newcommand{\dnum}{\delta^{\mathrm{num}}_{t}}
\newcommand{\rnum}{\rho^{\mathrm{num}}_{t}}
\usepackage{multirow}
\usepackage{pifont}
\usepackage{enumitem}
\usepackage{wrapfig}
\title{\minite: Diagnosing and Bridging the Execution Gap in Native NVFP4 Reinforcement Learning}
\author{%
}
\affiliation{\textbf{InfiX.ai}\quad \textbf{NVIDIA} \quad}
\code{\hypersetup{urlcolor=nv}\url{https://github.com/InfiXAI/miniTE}}
\logos{\raisebox{-0.5\height}{\includegraphics[height=6.5mm]{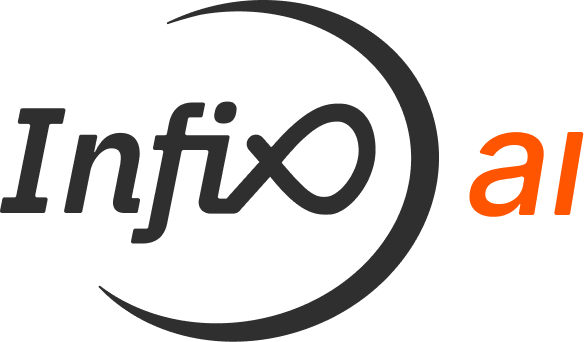}}%
\hspace{3mm}%
\raisebox{-0.5\height}{\includegraphics[height=3.5mm]{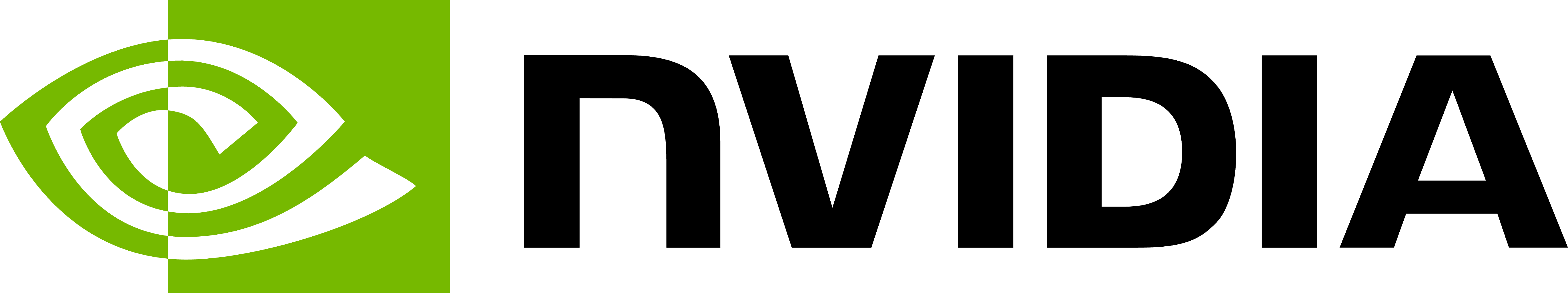}}}
\date{September, 2026}
\abstract{%
Native NVFP4 promises to accelerate reinforcement learning (RL) for large language models (LLMs), but achieving stable learning across separate training and sampling engines remains challenging. We show that matching model weights and numerical formats is insufficient to ensure consistent policy evaluation, highlighting the importance of cross-engine numerical execution. We introduce miniTE, a lightweight framework that combines execution diagnostics, cross-engine numerical alignment, and  efficient native W4A4 (4-bit weights and activations) GEMM computation for LLM RL. It improves numerical execution consistency without modifying the RL objective, while reducing implementation overhead to realize the benefits of native low-bit computation. Long-horizon RL experiments on Qwen3-30B-A3B demonstrate stable training and benchmark performance close to BF16, outperforming naive Transformer Engine (TE) NVFP4 in evaluation quality. On single NVIDIA B300 GPU node, the optimized learner reaches 33,540 tokens/s, yielding 1.39$\times$ and 1.47$\times$ the throughput of BF16 and TE NVFP4, respectively, for complete learner updates. We released the framework, kernels, diagnostics, and training configurations to make native low-bit RL easier.
}
\begin{document}

\maketitle

\section{Introduction}
\label{sec:intro}
\vspace{-2mm}

Reinforcement learning is central to large language model (LLM)
post-training, supporting instruction following \citep{ouyang2022training}
and mathematical reasoning \citep{shao2024deepseekmath,deepseekai2025r1}.
Repeated generation, policy evaluation, and updates \citep{yu2025dapo}
span an autoregressive sampler and a learner that optimizes on sampled
sequences. Native low-precision computation must accelerate these engines
while preserving effective learning.

Four-bit weight storage \citep{dettmers2023qlora} and native four-bit
training \citep{castro2025quartet,nvidia2025nvfp4} address different
execution paths. Quantized sampling, matched precision, and
quantization-aware RL address learner--sampler mismatch
\citep{qerl2025qerl,xi2026jetrl,yu2026quads,gu2026qarl}; recent systems
also support per-token NVFP4 and high-precision backward
\citep{cursor2026composer2,miles2026nvfp4rl}. We focus on native
weight-and-activation four-bit (W4A4) expert computation, connecting
cross-engine diagnosis, a practical numerical path, and complete learner gains.

RL makes cross-engine numerical behavior consequential. Probabilities
computed by the learner and sampler enter importance ratios and clipping
decisions \citep{schulman2017ppo,zheng2025gspo}; implementation differences
therefore enter optimization even at a shared parameter version.
Training--inference mismatch and execution-aligned references have been
studied in RL pipelines \citep{yao2025offpolicy,zhong2026vexact}.
NVFP4 adds choices in scale domains, rounding, accumulation, and operator
composition that a format name does not specify. This raises a concrete
question: \emph{which numerical dependencies must be exposed and controlled
when native low-bit computation spans two execution engines?}

Mixture-of-experts (MoE) models combine continuous expert computation
with discrete routing. Selective router precision and routing replay
address sensitivity in expert selection \citep{fedus2022switch,ma2025routing}.
We examine what remains after controlling parameter updates and
matching assignments at the observed layer. Frozen-policy scoring
shows that NVFP4 enlarges the cross-engine log-probability tail.
Target-layer routing agreement leaves FC1/FC2 output discrepancies,
and a sampler-only scaling change does not establish agreement
(Figure~\ref{fig:gap-diagnosis}). These controls identify an execution
specification, extending beyond bit width and expert IDs, as a useful
unit of diagnosis and implementation.

Motivated by these findings, we introduce \minite{} to connect this diagnosis
with practical native NVFP4 training (Figure~\ref{fig:framework}). Our framework checks quantized
operands, expert outputs, and policy evaluation across the sampler and
learner. Strict execution provides a cross-engine reference for diagnosis;
the default path combines our per-token quantization and native grouped-GEMM
implementation with BF16 backward. Its purpose is stable, high-throughput
learning under the existing RL objective. Arithmetic throughput is only
one component of this cost: routed expert workloads also require operand
preparation, representation conversion, and dispatch. We therefore
optimize quantization and grouped GEMM as a connected dataflow,
reducing repeated preparation around the numerical core so that FP4
arithmetic gains translate into faster learner updates. Comparisons between
our initial and optimized operator implementations check the numerical
effects of those changes. 

Long-horizon Qwen3-30B-A3B runs show sustained learning with \minite{}
and higher mathematical evaluation scores than naive NVFP4.
Same-recipe module comparisons show $2.46\times$ expert-forward speedup
over TE U3045. Complete learner updates on fixed text reach $1.39\times$
BF16 throughput in the optimized configuration and $1.33\times$ in a
repeated core study. The module comparison measures native implementation
gains; complete learner updates measure the numerical path and
implementation together.
Separate colocated RL measurements identify weight conversion and
synchronization as an integration bottleneck.
Our contributions are:
\begin{enumerate}[wide=8pt, leftmargin=8pt]
\item \textbf{Execution-gap diagnosis.} Frozen-policy, routing-conditioned,
and scaling controls identify numerical differences beyond nominal NVFP4
recipe agreement; strict execution supplies an actionable alignment
reference.
\item \textbf{Practical native NVFP4 RL.} An integrated W4A4 expert-forward
path with BF16 backward sustains long-horizon learning and improves
all four mathematics benchmarks over naive NVFP4 without modifying
the RL objective.
\item \textbf{A lightweight framework with numerically faithful
acceleration.} \minite{} exposes configurable numerical recipes, replaceable expert
operators, and reusable diagnostics for developing custom low-bit
training settings. Operator-level numerical checks
and complete learner benchmarks substantiate its execution
optimizations. We release Dockerfiles, reference configurations, and
single-command launch scripts to support reproducible experiments
and extensions across the supported low-bit RL paths.
\end{enumerate}

\section{Background and Problem Setup}
\label{sec:background}

\paragraph{Two executions of a policy.}
Let $\pi^L_\theta$ and $\pi^R_{\bar\theta}$ denote learner and sampler
policies, with $\bar\theta$ the sampled weight version.
Our non-strict RL runs use truncated importance sampling (TIS) to correct learner--sampler likelihood
differences. A corrected clipped policy-gradient term has the schematic form
\begin{equation}
\mathcal L(\theta)=\mathbb E_t\!\left[w_t^{\rm TIS}
\min\!\left(\rho_t(\theta)\hat A_t,
\operatorname{clip}(\rho_t(\theta),1-\epsilon_{\rm lo},1+\epsilon_{\rm hi})\hat A_t\right)\right],
\quad
\rho_t(\theta)=\frac{\pi^L_\theta(a_t\mid s_t)}{\pi^L_{\bar\theta}(a_t\mid s_t)},
\label{eq:clip}
\end{equation}
where $s_t$ is the token prefix, $a_t$ the chosen token, and $\hat A_t$ its
advantage \citep{schulman2017ppo,shao2024deepseekmath,yu2025dapo}.
The bounded correction $w_t^{\rm TIS}$ is computed from pre-update
learner-to-sampler likelihood ratios under the configured aggregation;
it is separate from clipping the learner's update ratio $\rho_t$.
Holding the parameter version fixed isolates the execution difference:
\begin{equation}
\dnum=\log\learner(a_t\mid s_t)-\log\rollout(a_t\mid s_t),
\qquad \rnum=\exp(\dnum).
\label{eq:delta-rho}
\end{equation}
We measure these quantities by teacher-forced scoring of identical
sequences. Holding the parameter version, token IDs, and scoring context
fixed separates execution differences from weight lag and sampling
variation.

\paragraph{What NVFP4 fixes.}
NVFP4 combines E2M1 values with one E4M3 scale per block of 16 values and
an outer FP32 scale \citep{nvidia2025nvfp4}. For a nonzero block $b$, outer
scale $g$, and block maximum $a_b=\max_{i\in b}|x_i|$, its reconstruction is
\begin{equation}
s_b=\operatorname{E4M3}\!\left(\frac{a_b}{6g}\right),\qquad
c_i=\operatorname{E2M1}\!\left(\frac{x_i}{g s_b}\right),\qquad
\hat x_i=g s_b c_i.
\label{eq:nvfp4}
\end{equation}
Here 6 is the largest E2M1 magnitude. If a rounded block scale is zero,
the implementation sets its codes to zero. The outer scale can cover an entire
tensor or a row. Row-wise scaling reduces dependence on maxima in other
rows and changes how block scales use the E4M3 dynamic range. It does not
fix rounding instructions, reduction order, padding, or where router
weights and output casts enter the computation. Those choices are part of
the execution specification.

\paragraph{Three comparisons.}
We distinguish quantization error (NVFP4 versus BF16 within a stack),
cross-stack execution difference (learner versus sampler), and
implementation difference (optimized versus reference code for the same
operator). Frozen-policy evaluation measures cross-stack execution difference;
reference-implementation tests measure the effect of an optimization;
training and benchmark evaluation measure learning quality. Throughout,
\emph{native W4A4} refers to the expert-forward GEMMs. Attention,
non-expert modules, backward, and optimizer state retain their specified
higher precisions.

\vspace{-3mm}
\section{Diagnosing the Execution Gap}
\label{sec:mechanism}

The frozen-policy study scores identical token sequences under teacher forcing with a frozen
Qwen3-30B-A3B-Base checkpoint \citep{yang2025qwen3}. Tails use 32 prompts
and 69,088 matched tokens; localization uses 12 prompts at 11 depths.
Confidence intervals use 2,000 prompt-cluster bootstrap resamples. Within-stack
self-comparisons are bitwise exact. Appendix~\ref{app:gap-extended}
specifies the collectors and full protocol.
\vspace{-2mm}
\subsection{Does the Gap Exist Before Optimization?}
\label{sec:mechanism-existence}

\begin{figure}[htbp]
    \centering
    \begin{subfigure}[t]{.45\linewidth}
        \centering
        \includegraphics[width=\linewidth]{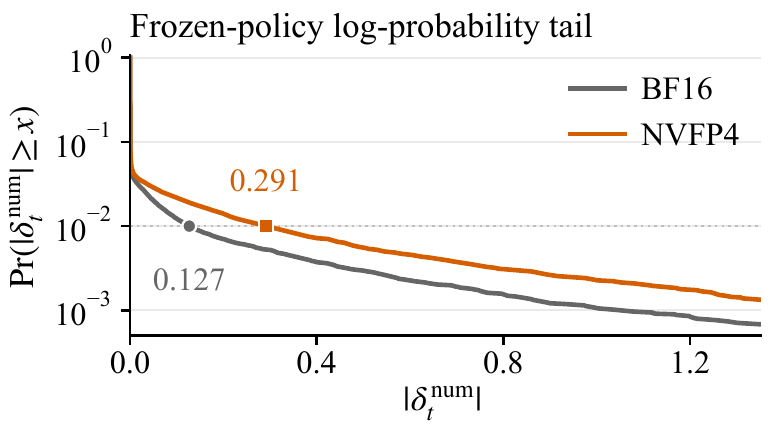}
        \label{fig:gap-survival}
    \end{subfigure}
    \hspace{.025\linewidth}
    \begin{subfigure}[t]{.45\linewidth}
        \centering
        \includegraphics[width=\linewidth]{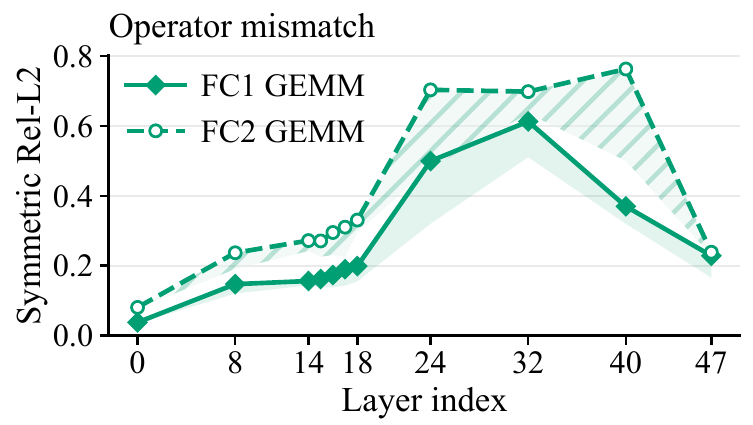}
        \label{fig:fc-discrepancy}
    \end{subfigure}
    \captionsetup{skip=2pt}
    \caption{\textbf{Execution differences survive recipe and routing agreement.}
    Left: Frozen-policy absolute logp tails, displayed through pooled P99.9;
    statistics use all matched tokens. Right: FC1/FC2 discrepancies with matched
    target-layer assignments at all 11 depths and no invalid experts;
    inputs retain upstream differences. Shading: 95\% prompt-cluster
    bootstrap intervals.}
    \label{fig:gap-diagnosis}
\end{figure}

The execution gap is already present at a shared, frozen parameter
version. The P99 of $|\dnum|$ increases from 0.1271 nats under BF16 to
0.2911 under NVFP4 (Figure~\ref{fig:gap-diagnosis}, left). The paired increase
is 0.1641, with a 95\% confidence interval of [0.1311, 0.2005]. NVFP4
therefore amplifies the log-probability tail beyond the BF16 cross-engine
floor before any optimizer update takes place.

Numerical-only ratios exceed the analysis clipping bounds for 1.203\%
of NVFP4 tokens versus 0.627\% under BF16. This is clipping potential;
training also depends on advantages, updates, and TIS correction.
Appendix~\ref{app:gap-extended} reports the full statistics.
\vspace{-2mm}
\subsection{Are Recipe and Routing Agreement Sufficient?}
\label{sec:mechanism-routing}
\label{sec:mechanism-granularity}

\paragraph{Matched routing leaves operator differences.}
We fix expert assignments at each target layer while retaining its
naturally accumulated inputs. FC1 symmetric relative-$L_2$ discrepancies
range from 0.0381 to 0.6124, and FC2 discrepancies from 0.0810 to 0.7624
(Figure~\ref{fig:gap-diagnosis}, right). Matched assignments therefore leave
residual output differences. Natural routing additionally shows changing
expert selections and nonmonotonic depth-dependent divergence.
A separate replay at a common MoE-layer input also leaves FC1/FC2
differences between learner modules and the sampler reference
(Appendix~\ref{app:common-input-replay}).

\paragraph{Finer scales alone do not establish consistency.}
Changing only the sampler activation outer scale from per-tensor to
per-token leaves FC1/FC2 discrepancies unchanged; the tail change is
inconclusive on this four-prompt cohort. The early per-token candidate
also retains a larger tail than standard NVFP4
(Appendix~\ref{app:gap-extended}). These controls motivate specifying
the complete operator path: finer scales alone do not establish
cross-engine agreement.

\paragraph{Implication for the training path.}
These findings motivate specifying the operands and composition that
define each execution. \minite{} makes these choices explicit and
implements a native training path, evaluated for learning behavior
and cost in Section~\ref{sec:experiments}.

\section{\minite{}: A Practical Native NVFP4 RL Framework}
\label{sec:method}

\begin{figure}[t]
\centering
\includegraphics[width=\linewidth]{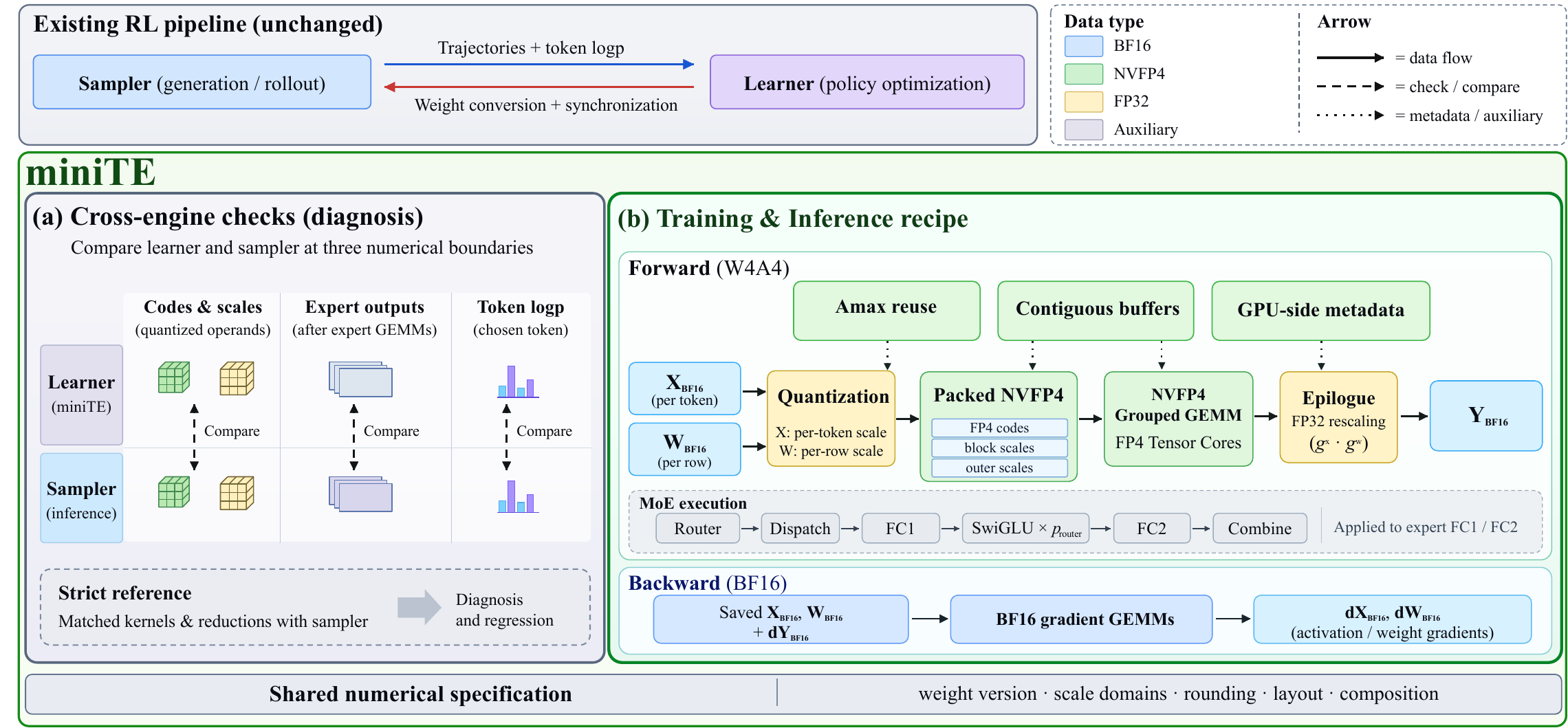}
\caption{\textbf{Numerical specification and efficient execution in \minite{}.}
(a) Cross-engine comparisons at operand, expert-output, and policy
boundaries, with a strict execution reference. (b) Packed W4A4 expert
forward, FP32 epilogue rescaling, and BF16 backward on saved
high-precision operands. Contiguous buffers, device metadata, and
producer statistics connect quantization directly to grouped GEMM.}
\label{fig:framework}
\end{figure}

Section~\ref{sec:mechanism} motivates making the executed numerical path
explicit. We choose a native recipe whose activation scales depend on
each token row, and jointly optimize the cost of executing it.
Figure~\ref{fig:framework} connects this numerical specification to
diagnostic interfaces and the native expert-computation dataflow.

\subsection{Execution Specification and Diagnostic References}
\label{sec:contract}

\paragraph{Operands and composition.}
An execution specification records scale domains, rounding, logical
layouts, and the placement of casts and router coefficients. Comparisons
fix the weight version, token IDs, and scoring context, then decode
codes and scales to a common layout. Input provenance and expert IDs
identify the comparison boundary: naturally accumulated inputs reveal
cross-engine differences, while common-input comparisons test the
operators applied at that boundary. This distinction follows directly
from the residual differences under matched target-layer routing.

\paragraph{Two complementary references.}
The strict execution reference matches learner kernels, formulas, and
reduction orders to the sampler, including surrounding normalization
and communication paths. It retains native NVFP4 expert forward and
BF16 gradients, and provides a cross-engine intervention evaluated in
Section~\ref{sec:determinism}. The operator reference instead compares
initial and optimized implementations on common inputs under a fixed
recipe. It tests whether an API or layout transformation preserves
quantized operands and outputs. The practical learner below is evaluated
as a complete learning configuration; its implementation checks use
the latter reference.

\subsection{The Native NVFP4 Learning Path}
\label{sec:qat}

\paragraph{Scaling with local activation dependence.}
Each dispatched activation row has its own outer scale, removing
dependence on other tokens' maxima under fixed quantization rules.
This addresses token-grouping differences between learner and sampler
at the quantizer boundary; weight scales, accumulation, and composition
complete the numerical path.
For expert $e$, token $t$, and weight output row $n$, the row-wise
native path uses the normal-range dequantization factors
\begin{equation}
 g^x_{e,t}=\frac{\max_k|X_{e,tk}|}{6\cdot448},\qquad
 g^w_{e,n}=\frac{\max_k|W_{e,nk}|}{6\cdot448}.
 \label{eq:expert-global-scales}
\end{equation}
Length-16 E4M3 block scales and E2M1 codes follow Eq.~(\ref{eq:nvfp4}).
Hadamard transforms and stochastic rounding are disabled in this path.
A context-swap test varies companion rows around fixed target rows.
The tested \minite{} path preserves target-row FP4 codes and GEMM
outputs bitwise; the RHT-enabled per-tensor reference changes in five
of six swaps.

\paragraph{Native accumulation and epilogue rescaling.}
The outer factors are independent of the reduction index $k$, so they
can be applied after the native block-scaled accumulation:
\begin{equation}
 U_e=\operatorname{MMA}_{\mathrm{NVFP4}}(\bar X_e,\bar W_e^{\mathsf T}),
 \qquad
 Y_{e,tn}=\operatorname{cast}_{\mathrm{BF16}}
 \!\left(g^x_{e,t}g^w_{e,n}U_{e,tn}\right).
 \label{eq:grouped-dataflow}
\end{equation}
Here $\bar X$ and $\bar W$ denote the packed FP4 operands and E4M3
block scales consumed directly by Tensor Cores. The epilogue broadcasts
the outer factors in FP32 before the BF16 output cast. This connects
the scaling rule to native arithmetic without materializing decoded
BF16 operands. Appendix~\ref{app:spec-correctness} specifies the floating-point
multiplication order and zero/tiny-row behavior.

\paragraph{Independent backward precision.}
The default backward uses a straight-through approximation with saved,
original BF16 activations and weights. For upstream gradient $G_e$,
BF16 GEMMs compute $\nabla X_e\approx G_eW_e$ and
$\nabla W_e\approx G_e^{\mathsf T}X_e$. This choice preserves high-precision
operands for the gradient path while avoiding backward quantization
and its preparation costs. Section~\ref{sec:exp-ablations} compares its
complete forward-plus-backward cost with the MXFP8 alternative.

\paragraph{MoE composition and RL integration.}
The learner applies router probabilities to the SwiGLU activation before
FC2, followed by expert combination. Padding, quantization, and casts
preserve this order. Non-expert modules retain their configured higher
precisions. Weight conversion and synchronization connect the learner
to an NVFP4 sampler with per-token activation scaling; the existing
TIS-corrected RL pipeline consumes their policy evaluations.

\subsection{Joint Design of Quantization and Grouped GEMM}
\label{sec:implementation}

Routed expert computation requires operand preparation and scheduling
in addition to matrix multiplication. Expert-dependent group sizes,
empty groups, and padding affect the interface between these operations.
We optimize that interface as a connected dataflow.

\paragraph{A shared producer--consumer representation.}
The reference quantizer already processes grouped inputs, but returns
expert views or lists from which grouped GEMM reconstructs its arguments.
\minite{} writes packed FP4 codes, block scales, and outer-scale information
directly into contiguous buffers consumed by GEMM. Device-side offsets
describe expert boundaries, and metadata storage and workspaces are reused.
This removes repeated splitting and argument preparation between two
grouped interfaces while retaining expert-dependent offsets and padding,
including support for empty groups.

\paragraph{Statistics travel with their activations.}
The FC1 reference permutes tokens, pads them, and scans the padded rows
for maxima. \minite{} collects maxima during permutation and pads both
the activations and statistics:
\[
\begin{array}{rcl}
\text{Reference}
&:&
\mathrm{sort}
\rightarrow
\mathrm{pad}(X)
\rightarrow
\boxed{\mathrm{amax\ scan}}
\rightarrow
\mathrm{quant}
\\[2mm]
\textsc{miniTE}
&:&
\boxed{\mathrm{sort}+\mathrm{amax}}
\rightarrow
\mathrm{pad}(X,\mathrm{amax})
\rightarrow
\mathrm{quant}
\end{array}
\]
FC2 obtains maxima from the router-weighted activation producer,
matching the tensor actually consumed by its quantizer. Joint handling of statistics and padding eliminates the extra activation scan without introducing a separate metadata-processing path. Producer-side statistics also appear in COAT's
group scaling \citep{xi2025coat}; here they cross MoE permutation,
padding, weighted activation, and NVFP4's two scale levels. Complete
learner timing includes the producer's work.

\paragraph{Layout and native GEMM execution.}
Fused scale layout preparation supplies the format consumed by Tensor
Cores. FC1 and FC2 use GEMM configurations adapted to their different
reduction lengths, coupling tile selection and tensor-memory allocation.
The optimized path combines these choices with the dense representation,
permutation fusion, and amax reuse. Sections~\ref{sec:exp-efficiency}
and~\ref{sec:exp-ablations} report implementation gains and numerical
checks for these transformations;
Appendix~\ref{app:results-extended} gives configuration details.

\paragraph{Framework integration.}
\label{sec:integration}
Users integrate \minite{} by replacing the expert grouped-linear module
through a common expert-computation interface. Quantization recipes and
backward precision are configurable separately, with BF16 and MXFP8
backward available for the supported NVFP4 paths. Compatible quantizer
and grouped-GEMM implementations provide layout and scheduling options
within the existing RL orchestration. The operand, expert-output, and
policy-evaluation boundaries in Section~\ref{sec:contract} provide reusable
comparison points for checking each recipe or kernel change.

\section{Experiments}
\label{sec:experiments}

We evaluate learning behavior, complete learner efficiency, and the
numerical and performance effects of the implementation design.
The frozen comparisons in Section~\ref{sec:mechanism} establish the
problem; the experiments here evaluate the training configurations
and execution paths.

\subsection{Experimental Setup}
\label{sec:exp-setup}

\paragraph{Model and configurations.}
We use Qwen3-30B-A3B-Base. The standard baseline
uses TE's per-tensor NVFP4 recipe on expert grouped linear layers
\citep{transformerengine}; non-expert learner modules remain BF16.
All non-strict RL arms use Truncated importance sampling (TIS). A separate strict
reference evaluates joint execution alignment without TIS.
Table~\ref{tab:rl-protocol} records the main RL and evaluation protocol;
Appendix~\ref{app:protocols} identifies the implementations.

\paragraph{Complete learner updates.}
Timing uses eight B300 GPUs, sequence length 4,096, and global batch
size 128: 524,288 tokens per update, including forward, backward,
communication, and optimizer work. All configurations share a BF16
checkpoint, seed, and frozen DAPO-Math-17k workload \citep{yu2025dapo}.
The core study uses ten process trials; the optimized study uses one,
with two warm-up updates followed by 15 measurements, trimmed to 13.
Appendix~\ref{app:protocols} specifies the workload and throughput estimators.

\paragraph{Module and routing controls.}
FC1 and FC2 use balanced expert workloads at four token counts, with
five process trials per cell and warm unchanged weights. Forward and
forward-plus-backward are measured directly. A separate eight-GB200
study crosses model initialization with input data to examine the
routed workload (Appendix~\ref{app:workload-sensitivity}).

\subsection{Long-Horizon Training Behavior and Evaluation}
\label{sec:exp-stability}

\begin{figure}[tbp]
\centering
\includegraphics[width=\linewidth]{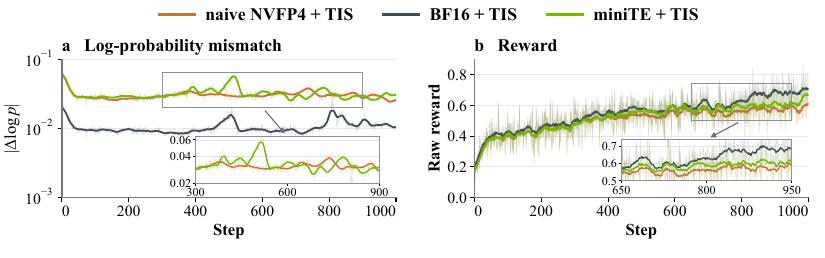}
\caption{\textbf{Learning with native NVFP4 and TIS.}
(a) Absolute learner--sampler logp difference; (b) reward. Each
configuration contributes one run. Faint curves show
raw values; solid curves show centered 21-step means. Insets magnify
marked regions.
Appendix~\ref{app:training-traces} gives endpoints and plotting details.}
\label{fig:longtrain}
\end{figure}

Figure~\ref{fig:longtrain} compares BF16, naive NVFP4+TIS, and \minite{}+TIS
through recorded log-probability discrepancy and reward
over the first 1,000 steps. \minite{} maintains reward growth
and achieves higher late-run reward than the standard NVFP4 trace,
while BF16 reaches higher reward over that region. The two NVFP4
paths have similar mean log-probability discrepancies despite their
different reward trajectories. Thus the observed learning behavior
is evaluated for the complete recipe; the mean training metric alone
does not identify which numerical differences affect learning.

\paragraph{Mathematics and instruction following.}
We prioritize mathematics because RL training uses DAPO math data.
At step 999 (Table~\ref{tab:quality}), \minite{} gains 10.00 and 3.88
percentage points over naive NVFP4 on AIME 2024 and 2025, followed by
gains of 0.55 on MATH-500 and 0.08 on GSM8K. The five-task mean rises
from 54.57\% to 56.35\%, remaining 2.49 points below BF16.
On IFEval \citep{zhou2023ifeval}, \minite{} scores 38.40\%, close to
BF16's 38.36\% but below NVFP4's 43.99\%, showing a task-dependent ranking.

\begin{table}[htbp]
\centering
\caption{\textbf{Evaluation results (\%).} Mean scores over
six evaluation rounds for each AIME benchmark and four for the others.
IFEval is Prompt-level Strict accuracy averaged over all four rounds.
Avg. equally weights the five task scores; $\Delta$ is its difference
from BF16 in percentage points. The gain row reports \minite{} minus NVFP4
for each task and the mean. BF16 uses BF16 inference; both NVFP4 paths
use NVFP4 inference. All three RL paths use TIS. Protocol details are in
Appendix~\ref{app:checkpoint-evaluation}.}
\label{tab:quality}
\small
\setlength{\tabcolsep}{3.5pt}
\begin{tabular}{@{}lrrrrrr@{}}
\toprule
Configuration & AIME24 & AIME25 & MATH-500 & GSM8K & IFEval & Avg. ($\Delta$) \\
\midrule
BF16 & 41.11 & 31.11 & 90.05 & 93.57 & 38.36 & 58.84 (0.00) \\
\midrule
NVFP4 & 26.67 & 20.56 & 86.90 & 94.71 & 43.99 & 54.57 ($-4.27$) \\
\minite{} & 36.67 & 24.44 & 87.45 & 94.79 & 38.40 & 56.35 ($-2.49$) \\
\cmidrule(l){2-7}
Gain vs. NVFP4 & $+10.00$ & $+3.88$ & $+0.55$ & $+0.08$ & $-5.59$ & $+1.78$ \\
\bottomrule
\end{tabular}
\end{table}

\subsection{Operator and Learner Efficiency}
\label{sec:exp-efficiency}

\paragraph{Complete updates.}
These measurements cover complete learner updates on fixed text,
rather than end-to-end RL iterations. The optimized \minite{} path
reaches 33,540 tokens/s, or
$1.39\times$ BF16 and $1.47\times$ TE NVFP4 throughput
(Table~\ref{tab:efficiency}). The repeated core study reaches 32,517
tokens/s and a trial-ID-paired trimmed speedup of $1.334\times$ over
BF16. The optimized setting combines the dense path, FC1/FC2 GEMM
configuration, permutation fusion, and activation-amax reuse. These
measurements establish the cost of that combined numerical and execution
configuration. The $1.334\times$ core estimate averages paired speedup
ratios after trimming; dividing its aggregate throughputs instead gives
$1.344\times$. The two estimators answer different aggregation questions
and are reported separately.

\begin{table}[htbp]
\centering
\caption{\textbf{Complete learner throughput on eight B300 GPUs.}
All fixed-text updates process 524,288 tokens; rollout and weight
synchronization are excluded. Optimized values are trimmed
within-run means from one process trial; core values aggregate ten
process trials. Core speedups use trimmed trial-ID-paired ratios,
whereas optimized speedups are same-round throughput ratios.}
\label{tab:efficiency}
\label{tab:rl-cost}
\small
\begin{tabular}{@{}llrrr@{}}
\toprule
Implementation & Configuration & Tokens/s & vs.\ BF16 & Trials \\
\midrule
Optimized & BF16 & 24,134 & $1.000\times$ & 1 \\
 & TE NVFP4 & 22,801 & $0.945\times$ & 1 \\
 & \minite{} & \textbf{33,540} & $\mathbf{1.390\times}$ & 1 \\
\midrule
Core & BF16 & 24,192 & $1.000\times$ & 10 \\
 & TE NVFP4 & 23,294 & $0.959\times$ & 10 \\
 & \minite{} & \textbf{32,517} & $\mathbf{1.334\times}$ & 10 \\
\bottomrule
\end{tabular}
\end{table}

\paragraph{Implementation gain under the same forward recipe.}
Relative to the pinned upstream per-token implementation (TE U3045),
\minite{} accelerates expert forward by a geometric mean of $2.46\times$
across FC1/FC2 and four workload sizes, ranging from $1.587\times$ to
$3.065\times$. This comparison measures the implementation and interface gains at the
expert-module boundary under a common forward recipe. Complete-update
comparisons measure the numerical recipe and surrounding execution together.
Module inputs are pre-permuted, dispatch is
excluded, and FC2 starts from an activation prepared outside the timer.
The module and learner gains therefore have different references and
timing boundaries. The full latency grid is in
Appendix~\ref{app:results-extended}.

\paragraph{Why realistic routed workloads matter.}
Balanced module timings do not predict the complete learner ranking:
\minite{} with BF16 backward is slower than BF16 in seven of the eight
module forward-plus-backward cells. Actual MoE execution additionally
involves irregular group sizes, preparation, and scheduling. In the
GB200 routing control, random weights with mock data activate 50.8
experts per layer on average, compared with 126.2 for pretrained weights
and real DAPO text, out of 128. The latter workload remains strongly
imbalanced across experts. The full $2\times2$ control in
Appendix~\ref{app:workload-sensitivity} establishes that initialization
and input data change the dispatched workload. It does not decompose
learner latency or assign the observed speedup to routing imbalance.
The control motivates using pretrained weights and real token sequences
for complete learner measurements; those measurements report the benefit
of the combined numerical path and implementation.

\paragraph{Costs beyond the learner.}
Standalone NVFP4 serving reaches 80,676 output tokens/s versus 49,119
for BF16 under fixed-length generation, a $1.64\times$ gain.
In a separate colocated RL window, learner time falls from 47.04 to
42.56 s, while weight conversion/synchronization rises from 9.94 to
55.26 s. The two measured stages sum to 56.98 s for BF16 and 97.82 s
for \minite{} (Table~\ref{tab:appendix-rl-components}), identifying a
weight-conversion and synchronization bottleneck. These sums omit
rollout and other iteration stages and therefore do not determine
whether a complete RL iteration is faster or slower. The serving
comparison uses fixed-length generation and supplies a separate measure
of sampling capacity. Appendix~\ref{app:results-extended} reports
these stage definitions and protocols.

\subsection{Execution References and Design Validation}
\label{sec:exp-ablations}

\begin{figure}[htbp]
\centering
\includegraphics[width=\linewidth]{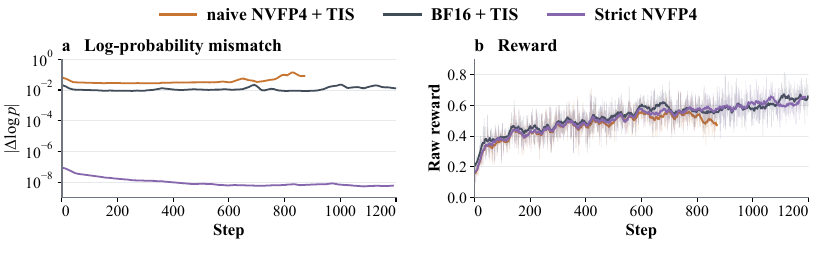}
\caption{\textbf{Strict execution in native NVFP4 RL.}
(a) Absolute learner--sampler logp difference; (b) reward, one run
per configuration. Faint curves show raw values; solid curves show
centered 21-step means.
Strict execution jointly changes operators and omits TIS.
Appendix~\ref{app:determinism} gives additional records.}
\label{fig:determinism}
\end{figure}

\paragraph{Strict execution reference.}
\label{sec:determinism}
The strict configuration matches learner kernels, formulas, and reduction
orders to the sampler and runs without TIS. It records a log-probability
difference of about $10^{-8}$ (Figure~\ref{fig:determinism}). This configuration
retains native NVFP4 expert forward and BF16 gradients, and maintains
reward through the region where the unaligned run deteriorates.
This provides an execution-level intervention in which native FP4
computation and stable learning coexist. The practical \minite{} path
is evaluated separately in Figure~\ref{fig:longtrain}.

\paragraph{Numerical preservation.}
The unfused quantization/grouped-GEMM API comparison passes all 28
bitwise checks against the upstream reference. A fused-versus-unfused
layout comparison gives zero deviation in the reported synthetic
and real-activation tests, including checks of externally computed
amax values. These results validate the tested operand and layout
transformations; Appendix~\ref{app:spec-correctness} gives their
shape and configuration coverage.

\paragraph{Backward precision and complete operator cost.}
Holding expert forward fixed, BF16 backward reduces directly measured
forward-plus-backward latency relative to MXFP8 backward in all eight
module cells, with a geometric-mean speedup of approximately
$1.21\times$. FP32-reference gradient comparisons appear in
Appendix~\ref{app:spec-correctness}.
The backward choice also has a measured numerical effect: across the
tested shapes and seeds, the largest gradient relative-$L_2$ difference
from the FP32 reference is 0.001663 for BF16 and 0.037653 for MXFP8.
Together, the latency and gradient comparisons support choosing
backward precision independently of the native W4A4 forward recipe.

\section{Related Work}
\label{sec:related}

\paragraph{Native low-precision learning.}
Microscaling specifies narrow elements and shared block scales
\citep{rouhani2023microscaling}. QLoRA uses four-bit storage with BF16
computation \citep{dettmers2023qlora}, whereas Quartet executes native
MXFP4 forward and gradient GEMMs \citep{castro2025quartet}.
NVFP4 pretraining \citep{nvidia2025nvfp4}, Composer~2
\citep{cursor2026composer2}, and Miles \citep{miles2026nvfp4rl}
provide native low-bit learning paths. Miles combines per-token NVFP4,
high-precision backward, and quantizer agreement. \minite{} studies
residual cross-engine execution differences and the implementation
cost of the native expert path.

\paragraph{Quantized RL.}
QeRL combines quantized rollout with low-rank adaptation
\citep{qerl2025qerl}; Jet-RL unifies FP8 training and rollout
\citep{xi2026jetrl}. QUADS combines learner-side W4A16 weight QAT
with sampler-side activation residual compensation \citep{yu2026quads}.
QaRL pairs low-bit forward computation with sequence-level trust-band
optimization \citep{gu2026qarl}. \minite{} evaluates native W4A4
expert-forward computation with BF16 backward and retains the existing
TIS-corrected RL pipeline.

\paragraph{Execution mismatch and routing.}
Off-policy analysis \citep{yao2025offpolicy}, sequence-level ratios
\citep{zheng2025gspo}, and execution-aligned references
\citep{zhong2026vexact} examine sampler--learner consistency.
FP16 studies address rounding precision \citep{qi2025fp16}; selective
precision \citep{fedus2022switch} and routing replay \citep{ma2025routing}
address expert selection. Deterministic kernels control dependence
on execution context
\citep{he2025nondeterminism,sglang2025deterministic,zhang2025tbik}.
Our diagnosis identifies residual differences under matched target-layer
assignments, while the native implementation is evaluated through
operator equivalence, learning traces, and complete learner timing.

\section{Conclusion}
\label{sec:conclusion}
Frozen-policy and routing-conditioned measurements expose execution
differences beyond nominal NVFP4 agreement. We introduced \minite{},
combining cross-engine diagnostics, native W4A4 forward with
independently selected backward precision, and jointly optimized
quantization and grouped GEMM. On Qwen3-30B-A3B, \minite{} sustained
learning over a long RL training horizon and outperformed standard
TE NVFP4 on all four mathematics benchmarks, with a five-task
average close to that of BF16. The optimized learner achieves
$1.47\times$ the throughput of TE NVFP4 and $1.39\times$ that of
BF16. Separate expert-forward measurements under the same recipe
show a $2.46\times$ speedup over Transformer Engine.
With configurable recipes, replaceable operators, and reproducible
launch environments, \minite{} provides an extensible foundation
for numerical and systems co-design in native low-bit RL.

\section{Limitations and Future Work}
\label{app:limitations}

\paragraph{Numerical fidelity beyond execution alignment.}
Although \minite{} improves mathematical evaluation over standard
TE NVFP4, its practical path still falls short of BF16 in average
evaluation quality and of strict execution in cross-engine
agreement. Matching execution does not eliminate the quantization
error associated with E2M1 values and E4M3 block scales.
However, the remaining quality gap does not isolate the effects
of quantization error from residual execution differences.
Future work should investigate tighter execution alignment
together with format-aware error compensation, aiming to
improve learning quality while retaining native W4A4 computation.

\paragraph{End-to-end efficiency and hardware-aware fusion.}
Our colocated measurements identify weight conversion and
synchronization as a substantial integration cost. These
measurements cover selected stages rather than complete RL
iterations, so component speedups alone do not establish
end-to-end gains. Reducing repeated conversion and packing
at learner--sampler boundaries is therefore a priority.
Online amax--quantization--GEMM fusion offers a complementary
direction for reducing operator-side preparation costs.
Its numerical behavior and performance need joint validation,
particularly when scales evolve during accumulation.
Efficient implementations must coordinate scale updates,
accumulator rescaling, and data movement within SM resource
constraints, motivating further hardware--software co-design.

Our learning evidence covers one MoE model with one run per
configuration, and the implementation gains are evaluated on
Blackwell hardware. Broader validation across models, tasks,
independent training seeds, and hardware platforms remains
future work.


\bibliography{iclr2027_conference}
\bibliographystyle{iclr2027_conference}
\clearpage
\appendix
\renewcommand{\topfraction}{0.9}
\renewcommand{\textfraction}{0.05}
\renewcommand{\floatpagefraction}{0.8}
\section{Contributors}
Zhen Li, Mingfa Feng, Yiming Zhang, Yang Yu, Changhai Zhou, Pengxiang Li, Dongkuan Wu, Congkai Xie, Hongxia Yang, Lin Yang, Jianmin Wu, Cheng Lin, Tao Zhao, Shuai Zhang, Shuang Yu, Yongming Shi,  Junjie Lai

\section{Experimental protocols and implementation identities}
\label{app:protocols}

\paragraph{TIS and experiment scope.}
All non-strict RL runs use truncated importance sampling (TIS),
including the BF16 control, naive NVFP4, and \minite{} paths.
The strict execution configuration runs without TIS. TIS reweights
policy-gradient contributions using bounded learner-to-rollout
likelihood ratios; it is distinct from policy-update ratio clipping
and from numerical execution alignment. The frozen-policy diagnostics,
isolated modules, and fixed-text learner benchmarks do not apply an
RL importance correction. Their numerical-format labels therefore
remain BF16, NVFP4, or \minite{} without a TIS suffix.

\paragraph{Model and execution environment.}
The learner benchmarks use Qwen3-30B-A3B, with 48 layers, hidden size
2,048, 128 experts, top-8 routing, and expert intermediate size 768.
The node contains eight NVIDIA B300 SXM6 GPUs with 275\,GB of memory
per GPU. The software stack uses PyTorch 2.11.0+cu130, CUDA 13.0.88,
NCCL 2.28.9, Megatron Core 0.16.0rc0, and driver 590.48.01.
Megatron-LM uses revision \texttt{52153937}. The core repeated
learner benchmark uses \minite{} revision \texttt{953f17f9}; the
optimized batch uses revision \texttt{b641179}, which includes the
dense grouped path and FC1/FC2-specific GEMM-kind selection.
The initial Transformer Engine
revision, \texttt{eaca54e9}, was updated to \texttt{8ea5a92d} to support
zero-token experts. Both revisions retain the same three integration
patches, and their nonempty-expert paths were checked for parity.
The upstream per-token reference uses revision \texttt{690ffea4} in a
separate container built from the same image, with a CUDA 13 compilation
compatibility patch.

\begin{table}[!htbp]
\centering
\caption{Experimental settings and their numerical scope. W4A4
describes the forward expert computation; backward precision is specified
separately. The upstream reference quantizes a broader layer set.}
\label{tab:appendix-identities}
\small
\begin{tabular}{@{}lp{0.68\linewidth}@{}}
\toprule
Setting & Implementation and numerical scope \\
\midrule
BF16 & TE BF16 \texttt{GroupedLinear}; BF16 forward and backward. \\
NVFP4 & TE per-tensor NVFP4 recipe on expert grouped linear layers;
nonexpert linear layers remain BF16 in this environment. \\
CUTLASS NVFP4 & Custom CUTLASS NVFP4 expert forward,
with randomized Hadamard transform enabled by default. \\
\minite{} & Per-token NVFP4 expert
forward, with Hadamard transforms and stochastic rounding disabled. \\
MXFP8 backward & Quantized dgrad and wgrad for CUTLASS NVFP4 or \minite{}. \\
BF16 backward & \texttt{DISABLE\_BACKWARD\_QUANT=1}; BF16 dgrad and
wgrad under the straight-through estimator. \\
TE U3045 & Upstream PR\,\#3045 per-token NVFP4 implementation; includes
NVFP4 backward and nonexpert quantization in its full-model run. \\
\bottomrule
\end{tabular}
\end{table}

\paragraph{Complete learner updates.}
Tensor, pipeline, context, and expert parallelism are respectively
1, 1, 1, and 4. Sequence length is 4,096, microbatch size is 1, and
global batch size is 128, giving 524,288 tokens per optimizer update
and 16 accumulation microbatches. All configurations start from the
same BF16 checkpoint and seed 20260913. The input is a frozen corpus
of 657 sequences of 4,097 tokens constructed from DAPO-Math-17k text,
with an approximately 4.3-fold repetition factor in the construction.
These are complete training updates on fixed text, with forward,
backward, optimizer, and data-loading costs, rather than complete RL
iterations. The interval timer synchronizes ranks and measures the
slowest-rank wall time.

The setup uses distributed optimization, grouped GEMM, all-to-all
expert dispatch, and an FP32 router. Activation recomputation, offload,
and CUDA graphs are disabled. Data indexing is cached across runs.
CUTLASS NVFP4 and \minite{} pad routed expert workloads to multiples of 64; the
associated padding cost is included. The initial repeated experiment
uses unfused MoE permutation. The optimized configuration enables
\texttt{--moe-permute-fusion} for all variants and additionally
\texttt{--nvfp4-pertoken-amax-fuse} for the \minite{} variants, together
with the dense path and GEMM-kind selection in revision \texttt{b641179}.

\paragraph{Timing estimators.}
For the repeated experiment, each independent process runs 20 updates.
Updates 1--5 are warmup, and updates 6--15 form the measured window.
Let $t_{c,j}$ be the median measured update time in seconds for
configuration $c$ and process trial $j$. Its throughput is
$r_{c,j}=524{,}288/t_{c,j}$. For a vector of ten trial values, define
\begin{equation}
 \operatorname{trim}_{10}(x)
 =\frac{1}{8}\sum_{j=2}^{9}x_{(j)},
 \label{eq:appendix-trim}
\end{equation}
where $x_{(j)}$ denotes the $j$th order statistic. We report throughput
as $\operatorname{trim}_{10}(r_c)$ and paired speedup as
$\operatorname{trim}_{10}(r_{c,j}/r_{\mathrm{BF16},j})$.
The latter differs from the ratio of the two aggregated throughputs.
For example, \minite{} gives 1.334$\times$ under the paired estimator and
1.344$\times$ under the ratio of aggregated throughputs.

The optimized experiment has one process trial per configuration.
It runs 17 updates, discards updates 1--2, and averages the remaining
15 per-update throughputs after removing one minimum and one maximum.
Its 13 retained updates describe within-run timing, not 13 independent
process trials. Optimized and repeated results are reported separately
because the fusion settings and estimators differ.

\paragraph{Run order and memory.}
The first three BF16, NVFP4, and CUTLASS NVFP4 trials used a rotated configuration
order. The first BF16-backward CUTLASS NVFP4 and \minite{} trials were collected later
and match these runs by trial ID rather than by simultaneous execution.
Trials 0--2 of BF16, NVFP4, and CUTLASS NVFP4 used the initial TE revision; the
remaining runs used the zero-token-expert fix. During trial 3, an idle
serving process occupied GPU resources. This trial is the minimum for
NVFP4 and all three CUTLASS NVFP4/\minite{} variants and is removed by the stated trimming
rule. All ten observations remain visible in Table~\ref{tab:appendix-trials}.
Memory values denote the maximum reported \emph{rank-0 reserved}
allocation, in MiB. They are neither allocated-tensor memory nor a
maximum across all ranks.

\section{Additional execution-gap diagnostics}
\label{app:gap-extended}

The frozen-policy diagnostic uses a Qwen3-30B-A3B base checkpoint at
step zero. Both engines receive identical token IDs and score the same
chosen-token sequence under teacher forcing. The tail cohort contains
32 prompts in length buckets 512, 2,048, and 4,096, yielding 69,088
matched tokens per condition. Confidence intervals use paired
prompt-cluster bootstrap with 2,000 resamples and seed 20260825.
Whole-network log probabilities are counted once per prompt; captures
at different layers do not supply independent copies of these observations.

\begin{table}[!htbp]
\centering
\caption{Frozen-policy tail diagnostics, including the early per-token
candidate. The numerical-only ratio uses the same model parameters in
both engines. Clipping potential is its fraction outside $[0.80,1.28]$.}
\label{tab:appendix-tail}
\small
\begin{tabular}{@{}lrrrr@{}}
\toprule
Condition & $|\Delta\log p|$ P99 & Ratio P01 & Ratio P99 & Clip potential \\
\midrule
BF16 & 0.1271 & 0.9471 & 1.0489 & 0.627\% \\
NVFP4 & 0.2911 & 0.8933 & 1.1243 & 1.203\% \\
Per-token candidate & 0.3179 & 0.8745 & 1.1152 & 1.282\% \\
\bottomrule
\end{tabular}
\end{table}

\paragraph{A per-token candidate does not close the gap.}
The early candidate pairs a \texttt{nvfp4\_cursor2} learner with
per-token NVFP4 rollout. Relative to standard NVFP4, its P99 absolute
log-probability discrepancy increases by 0.0268 nats
(95\% CI $[0.0017,0.0495]$), and its clipping potential increases by
0.080 percentage points (95\% CI $[0.005,0.163]$). This candidate
changes more than rollout scaling and has a different identity from the
\minite{} learner benchmark. Its result motivates an execution specification
that covers the full operator path.

\paragraph{Depth and routing controls.}
Localization uses 12 prompts, with four each at lengths 256, 512, and
1,024, across layers $0,8,14,15,16,17,18,24,32,40,47$.
The three conditions yield 33 layer runs. Natural routing disagreement
ranges from 0.0459 to 0.5938 under BF16 and from 0.0459 to 0.7559
under NVFP4. MoE-output relative $L_2$ discrepancy ranges from
0.0072 to 0.2072 and from 0.0269 to 0.2831, respectively.
The routing-conditioned comparison fixes expert assignments at the
observed layer, where set-match is 1.0 and no invalid experts occur.
FC1 discrepancy remains 0.0381--0.6124, peaking at layer 32;
FC2 discrepancy remains 0.0810--0.7624, peaking at layer 40.
Upstream routing is allowed to differ, so this intervention establishes
residual mismatch conditional on target-layer assignments rather than
all-layer fixed-path propagation. Local contractions in the curves
show that divergence depends nonmonotonically on depth.

\begin{figure}[!htbp]
\centering
\includegraphics[width=\linewidth]{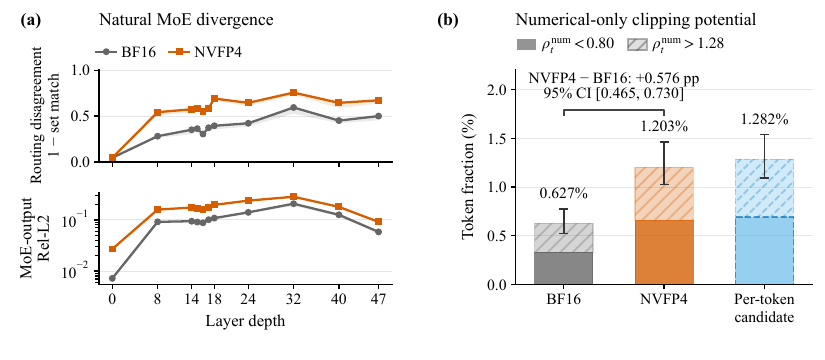}
\caption{\textbf{Natural routing and numerical clipping potential.}
(a) Routing disagreement and MoE-output discrepancy vary with depth
under natural execution. (b) The numerical-only ratio falls outside
$[0.80,1.28]$ more often under NVFP4 than under BF16.
This is a frozen-policy sensitivity diagnostic, not the clip fraction
recorded during an RL optimization step. Shaded bands and error bars show
95\% prompt-cluster bootstrap intervals; the per-token candidate is the
early diagnostic configuration.}
\label{fig:appendix-natural-clip}
\end{figure}

\begin{table}[!htbp]
\centering
\caption{Rollout-only scaling control on a separate four-prompt cohort.
The learner, checkpoint, inputs, and observed layer are fixed.}
\label{tab:appendix-q5}
\small
\begin{tabular}{@{}lrrr@{}}
\toprule
Metric & Per-tensor rollout & Per-token rollout & Difference \\
\midrule
FC1 GEMM relative $L_2$ & 0.03767 & 0.03767 & 0.00000 \\
FC2 GEMM relative $L_2$ & 0.08064 & 0.08064 & 0.00000 \\
MoE output relative $L_2$ & 0.02715 & 0.02362 & $-0.00353$ \\
$|\Delta\log p|$ P99 & 0.7596 & 0.7239 & $-0.0357$ \\
Clipping potential & 5.486\% & 5.329\% & $-0.157$ pp \\
\bottomrule
\end{tabular}
\end{table}

Changing rollout global scaling alone leaves FC1 and FC2 discrepancy
unchanged (Table~\ref{tab:appendix-q5}). The P99 difference has a 95\%
CI of $[-0.2231,0.1517]$, and the clipping-potential difference has
a CI of $[-0.783,1.309]$ percentage points. These intervals include
zero. The absolute tail values belong to this four-prompt cohort and
are not directly comparable to the 32-prompt values in
Table~\ref{tab:appendix-tail}.

\paragraph{Common-input replay.}
\label{app:common-input-replay}
A separate offline replay uses the same 12-prompt localization cohort
at layers 0, 24, and 47, with four prompts at each length of 256, 512,
and 1,024 tokens. It uses the Qwen3-30B-A3B base checkpoint on B300,
with \minite{} revision \texttt{b641179}, Transformer Engine
\texttt{8ea5a92d}, Megatron-LM \texttt{52153937}, and SGLang
\texttt{12344e4d}. The \minite{} learner uses EP2, list packing, and
expert padding for both capture and replay. The sampler uses one GPU,
TP1, and the Triton backend, loading BF16 or ModelOpt NVFP4 HF weights
for the corresponding path, with per-token activation scaling for NVFP4;
the learner loads the BF16 checkpoint.
Learner replay runs in evaluation mode with dropout disabled and no
gradient recording. Learner and sampler initialization seeds are 1234
and 20260824, respectively, without per-cell RNG-state restoration.
These results belong to this replay environment;
they are separate from the earlier frozen-policy tail measurements.

Inputs $h_L$ and $h_R$ are captured at the MoE-module entry, before
routing. We evaluate the four
combinations $F_L(h_L)$, $F_L(h_R)$, $F_R(h_L)$, and $F_R(h_R)$,
holding learner-captured expert assignments and each path's weights fixed.
Learner FC1/FC2 outputs are obtained by hooks on the actual replayed
MoE module. Sampler per-expert FC1/FC2 outputs are
\texttt{replayed\_reference}; its combined output is obtained from the
\texttt{actual\_kernel}. For NVFP4, the sampler reference
quantizes/dequantizes activations and dequantizes stored weights before
FP32-typed PyTorch matrix multiplication and BF16 output casts; the BF16 reference
uses unquantized operands.
Table~\ref{tab:common-input-replay} uses the
per-expert outputs, not hidden intermediates captured from the
sampler's fused production kernel. The sampler renormalizes captured
router coefficients per token, while the learner uses them directly.
Thus these comparisons retain the paths' composition rules.

Let $F_s^j(h)$ denote the concatenated expert outputs at FC$j$ for
path $s$. For each prompt, outputs are concatenated in expert-index
order and compared in FP32 using
\begin{equation}
 R(A,B)=\frac{\lVert A-B\rVert_F}{\lVert B\rVert_F},\qquad
 I_j=R\bigl(F_L^j(h_L),F_L^j(h_R)\bigr),\qquad
 E_j=R\bigl(F_L^j(h_R),F_R^j(h_R)\bigr).
 \label{eq:common-input-replay}
\end{equation}
Here $I_j$ measures an input swap within the learner path, and $E_j$
compares the two paths at $h_R$. The input discrepancy is $R(h_L,h_R)$.
A zero denominator returns zero for identical zero tensors and infinity
otherwise. We report the arithmetic mean of the 12 prompt-level values.
This reference-normalized metric differs from the symmetric relative-$L_2$
metric used in the right panel of Figure~\ref{fig:gap-diagnosis}; the columns also have
different denominators and are not additive source contributions.

\begin{table}[!htbp]
\centering
\caption{Common-input replay at three depths. Values are means over
12 prompts. $I_j$ is the learner input-swap discrepancy; $E_j$ compares
learner module outputs with the sampler reference at a common MoE-layer
input. FC2 operands are not held equal.}
\label{tab:common-input-replay}
\small
\setlength{\tabcolsep}{5pt}
\begin{tabular}{@{}llrrrrr@{}}
\toprule
& & & \multicolumn{2}{c}{FC1} & \multicolumn{2}{c}{FC2} \\
\cmidrule(lr){4-5}\cmidrule(l){6-7}
Setting & Layer & $R(h_L,h_R)$ & $I_1$ & $E_1$ & $I_2$ & $E_2$ \\
\midrule
BF16 & 0 & 0.1222 & 0.1231 & 0.1606 & 0.1103 & 0.1609 \\
& 24 & 0.7353 & 0.6049 & 0.1264 & 0.8211 & 0.2276 \\
& 47 & 0.4666 & 0.3425 & 0.1090 & 0.8957 & 0.2402 \\
\midrule
\minite{} & 0 & 0.0030 & 0.0220 & 0.0575 & 0.0383 & 0.0931 \\
& 24 & 0.2603 & 0.2349 & 0.0664 & 0.3895 & 0.1589 \\
& 47 & 0.1632 & 0.1501 & 0.0587 & 0.1541 & 0.0870 \\
\bottomrule
\end{tabular}
\end{table}

At layer 0, \minite{} retains discrepancies of 0.0575 at FC1 and
0.0931 at FC2 under the common input. The common input is the MoE-layer
input, so the FC2 comparison includes preceding FC1, activation, and
composition differences rather than isolating FC2 GEMM arithmetic.
The replay establishes residual output differences at these module and
reference boundaries; it does not partition the natural discrepancy
into causal percentages.

\section{Numerical specification and correctness tests}
\label{app:spec-correctness}

\paragraph{What an execution specification records.}
An executable numerical specification includes the quantized operands,
the inner block and outer scale domains, scale dtypes and rounding,
Hadamard and stochastic-rounding settings, logical layouts and padding,
expert routing and packing, accumulation and epilogue precision, and
the nonquantized operators surrounding the GEMM. Matching an FP4
encoding alone leaves these choices open. In \minite{}, activations use
length-16 blocks with E4M3 block scales and per-row FP32 outer scaling.
The executed grouped path quantizes each weight output row with the
same length-16 blocks and a separate FP32 outer scale per output row.
The encoder clamps its row maximum as $\max(a,10^{-12})$ before
forming the global encode scale. A block whose decoded E4M3 scale
is zero uses a zero encoding multiplier. The CUTLASS epilogue applies
the outer scales in the order
\begin{equation}
 Y_{mn}=\operatorname{BF16}\!\left[
 2688^{-2}\,a^{W}_{n}\bigl(a^{X}_{m}U_{mn}\bigr)\right],
 \label{eq:appendix-epilogue}
\end{equation}
where $U$ is the FP32 accumulator after inner-block dequantization,
and $a^{X}$ and $a^{W}$ are the activation-row and weight-row maxima.
The intermediate scale products stay in FP32 before the final BF16
rounding. The BF16-backward variant applies the straight-through
estimator to the unquantized saved operands.

\paragraph{Context sensitivity of the native path.}
\label{app:context-swap}
The context-swap test holds eight target activation rows fixed while
changing the companion rows in a 128-row FC1 input. BF16 sampler
activations from three prompts, of lengths 256, 512, and 1,024, are
captured at layer 0, expert 79. The input and weight shapes are $128\times2048$ and
$1536\times2048$; target outputs have shape $8\times1536$.
Within each case, target rows are the middle eight rows after sorting
that prompt's rows by maximum magnitude. Original companions come
from the same prompt. Low- and high-amplitude replacements are selected
from a pool of 2,955 captured rows across all 12 prompts. The current prompt is
excluded when selecting the lowest- and highest-amplitude companions.
The 256-token case has
96 companion rows and 24 zero-padding rows; the other two have
120 companions. Target positions, matrix shape, padding, and expert
weights remain fixed across contexts.

The per-tensor NVFP4 reference enables RHT; \minite{} disables it.
The reference uses a fixed $16\times16$ Hadamard transform and a
fixed sign mask, reused across contexts. Stochastic rounding is
disabled in both paths. The script does not reset the RNG between
contexts; the transform is fixed by the implementation and randomly
initialized module weights are overwritten with the same checkpoint
expert weights. Each path therefore holds its own numerical rules
fixed during the swap, although the two paths differ in RHT as well
as scaling and are not a single-variable scaling ablation.

\begin{table}[!htbp]
\centering
\caption{Context-swap results at layer 0, expert 79. $a$ is the shared
amax recorded by the per-tensor NVFP4 path. $\Delta_q$ and $\Delta_Y$
are maximum absolute changes in target-row dequantized activations and
native GEMM outputs, computed by subtracting BF16 tensors relative
to the original context; dashes mark that
reference. These are within-path context differences, not errors
relative to BF16. Dequantized values are compared within each path's
coordinates: after RHT for NVFP4, and without RHT for \minite{}.}
\label{tab:context-swap}
\small
\setlength{\tabcolsep}{5pt}
\begin{tabular}{@{}llrrrrr@{}}
\toprule
& & & \multicolumn{2}{c}{NVFP4 (per-tensor)} & \multicolumn{2}{c}{\minite{}} \\
\cmidrule(lr){4-5}\cmidrule(l){6-7}
Prompt length & Context & $a$ & $\Delta_q$ & $\Delta_Y$ & $\Delta_q$ & $\Delta_Y$ \\
\midrule
256 & Original & 3.156250 & --- & --- & --- & --- \\
& Low & 2.656250 & 0.217773 & 0.054932 & 0 & 0 \\
& High & 3.171875 & 0.203125 & 0.035156 & 0 & 0 \\
\midrule
512 & Original & 3.156250 & --- & --- & --- & --- \\
& Low & 2.671875 & 0.219727 & 0.051758 & 0 & 0 \\
& High & 3.171875 & 0.167969 & 0.031250 & 0 & 0 \\
\midrule
1,024 & Original & 3.171875 & --- & --- & --- & --- \\
& Low & 2.687500 & 0.164062 & 0.046143 & 0 & 0 \\
& High & 3.171875 & 0 & 0 & 0 & 0 \\
\bottomrule
\end{tabular}
\end{table}

Across all six swaps, \minite{} preserves the target FP4 payload
hashes and passes bitwise equality checks on the target GEMM outputs;
the dequantized target values and logged row-wise outer factors also
remain unchanged.
The \minite{} block-scale tensors are not separately compared in this test.
Five swaps change the per-tensor shared amax and its target outputs.
Its payload and block-scale hashes change in those same five swaps.
The high-amplitude context in the 1,024-token case leaves that amax
and the outputs unchanged, providing a negative control.
These observations validate context independence for the tested
\minite{} operator path and contexts. They do not measure quantization
accuracy against BF16, whole-network batch invariance, or RL stability.

\paragraph{Forward API equivalence.}
The upstream-reference test compares quantized payloads, decoded
scales, outer scales, and GEMM outputs under a common input. The
28 comparisons comprise 12 activation checks, 12 weight checks, and
four GEMM checks. All pass bitwise for $K=2048$, $N=1536$, and expert
splits $[128,256,512,128]$. The fork is configured with
\texttt{with\_swizzle=False}, \texttt{do\_amax=True},
\texttt{with\_rht=False}, and the default GEMM kind.
This establishes forward equivalence for the tested API paths and
shapes. Backward and fused-swizzle paths have separate tests.

\paragraph{Fused scale layout.}
The \minite{} per-token path compares fused and unfused scale layouts in
six synthetic cases with 32 experts and one real-activation case
with splits $[128,256,512,128]$. Synthetic inputs use $K=2048$,
$N=1536$, and 256, 1,024, 4,096, or 8,192 rows per expert; the
256-row case uses three seeds. All seven cases have identical FP4
payloads, row maxima, and decoded scales after restoring a common
layout, and bitwise-identical BF16 GEMM outputs. Internal amax
calculation and externally computed row maxima also agree bitwise
on these inputs. These tests cover the per-token API with
128-aligned splits, disabled Hadamard transforms, and the default
GEMM kind.

\begin{table}[!htbp]
\centering
\caption{Backward comparisons to FP32 gradient references, using
straight-through-estimator semantics. Values are the worst reported
across the tested seeds and shapes; CUTLASS NVFP4 and \minite{} have the same reported
bounds. BF16 torch matmul attains the same BF16-backward values.}
\label{tab:appendix-gradient}
\small
\begin{tabular}{@{}llrrr@{}}
\toprule
Backward & Gradient & Cosine, min. & Relative $L_2$, max. & Sign agreement, min. \\
\midrule
MXFP8 & dgrad & 0.999292 & 0.037626 & 98.802\% \\
MXFP8 & wgrad & 0.999291 & 0.037653 & 98.802\% \\
BF16 & dgrad & 0.999999 & 0.001660 & 100.000\% \\
BF16 & wgrad & 0.999999 & 0.001663 & 100.000\% \\
\bottomrule
\end{tabular}
\end{table}

\paragraph{Extreme inputs and gradient accuracy.}
The numerical checks include 24 extreme-input cases across activation,
weight, and dense per-token quantization, covering magnitudes near
$10^{30}$, the BF16 maximum, subnormals, and mixed magnitudes.
The observed boundary behavior is saturation or underflow rather
than NaN/Inf propagation. Subnormal handling depends on the path:
some activation scales flush to zero, whereas a weight path retains
FP32 subnormal scales and uses a 3\% scale tolerance in that regime.
This behavior follows from the represented dynamic range and is
distinct from layout/API equivalence. The backward checks in
Table~\ref{tab:appendix-gradient} establish high directional agreement
with FP32 references. They do not measure final policy quality.

\section{Training-trajectory provenance and strict alignment}
\label{app:determinism}

\subsection{Main RL protocol and training traces}
\label{app:training-traces}

\begin{table}[!htbp]
\centering
\caption{\textbf{RL training and evaluation settings.} Training uses our
launch configuration with eight nodes and expert parallelism of four.
Checkpoint selection and repetitions follow the evaluation records.}
\label{tab:rl-protocol}
\small
\setlength{\tabcolsep}{4pt}
\begin{tabular}{@{}>{\raggedright\arraybackslash}p{.28\linewidth}>{\raggedright\arraybackslash}p{.68\linewidth}@{}}
\toprule
Field & Setting \\
\midrule
\multicolumn{2}{@{}l}{\emph{Main training}} \\
Initialization & Qwen3-30B-A3B-Base actor checkpoint in \texttt{torch\_dist} format; BF16 or NVFP4 HF weights initialize the corresponding sampler. Existing experiment checkpoints are used on resume. \\
Seeds & Training: 1234; rollout shuffle: 42. \\
Data / algorithm & DAPO-Math-17k, chat template, shuffled prompts, math reward; GRPO. \\
Optimizer & Adam with decoupled weight decay 0.1; $\beta_1=0.9$, $\beta_2=0.98$, $\epsilon=10^{-8}$; gradient-norm clipping 1.0. \\
Learning rate / schedule & $10^{-6}$, constant; no warmup. \\
Batch / GRPO group & 32 prompts per rollout batch, 16 responses per prompt; global batch 512 responses. Dynamic microbatching with a 16,384-token/GPU budget. \\
Rollout generation & Temperature 1.0, top-$p$ 1.0, top-$k$ disabled; maximum response length 20,480 tokens. No explicit prompt/context limit is set. \\
GPU allocation & 8 nodes with 8 actor GPUs per node (64 total); colocated actor/sampler; 1 GPU per sampler engine. \\
Parallelism & TP/PP/CP/EP/ETP $=1/1/1/4/1$. \\
Memory / optimizer execution & Full activation recomputation, uniform one-layer checkpoints; distributed precision-aware optimizer with CPU offload and overlapped transfers. \\
Sampler backend & BF16: Triton. NVFP4: FlashInfer TRT-LLM with ModelOpt FP4 and per-token activation scaling. \\
\midrule
\multicolumn{2}{@{}l}{\emph{Correction}} \\
Policy-ratio clipping & $[0.80,1.28]$. \\
TIS & Enabled for non-strict RL; token-wise $w_t=\operatorname{clip}(\exp(\log p^L_{\rm old}-\log p^R),0,2)$, applied to the policy-gradient term after policy-ratio clipping. \\
Routing replay & Disabled. \\
\midrule
\multicolumn{2}{@{}l}{\emph{Evaluation}} \\
Checkpoint / precision & Step 999; BF16 inference for the BF16 checkpoint, NVFP4 inference for both NVFP4 paths. \\
Rounds / IFEval metric & Six rounds per AIME benchmark; four for MATH-500, GSM8K, and IFEval. IFEval uses mean Prompt-level Strict accuracy. \\
Decoding & Temperature 0.7, top-$p$ 1.0, maximum generation length 20,480 tokens. \\
\bottomrule
\end{tabular}
\par\smallskip
\parbox{\linewidth}{\footnotesize
The three main runs share the optimization and sampling settings above;
their numerical paths differ as described in Appendix~\ref{app:protocols}.
TIS uses token-wise importance weights; sequence averaging of the loss
is a separate reduction.}
\end{table}

The main training figure uses one exported run for each of BF16,
naive NVFP4, and \minite{}, all with TIS. Its panels use the logged metrics
\path{train/train_rollout_logprob_abs_diff} and \path{rollout/raw_reward},
respectively. It compares the common interval from
step 0 through step 999. The plot retains the raw observations and overlays a centered
21-point arithmetic moving average, with shorter windows at the
endpoints. The interval is selected before smoothing. The figure shows
raw traces and their moving averages without exported min--max bands.
The log-probability
inset shows steps 300--900 and values 0.02--0.065 on a logarithmic
axis; the reward inset shows steps 650--950 and values 0.50--0.74.
Both insets reuse the same observations and smoothed curves as their
parent panels.


\subsection{Checkpoint evaluation and instruction following}
\label{app:checkpoint-evaluation}

\paragraph{Protocol and configuration identity.}
Table~\ref{tab:quality} evaluates the step-999 checkpoints from the
BF16, standard NVFP4, and \minite{} training runs used in Figure~\ref{fig:longtrain}.
The BF16 checkpoint uses BF16 inference; the standard NVFP4 and
\minite{} checkpoints use NVFP4 inference. The row labels describe
TIS-corrected RL configurations. Thus the BF16 comparison covers both
training and evaluation paths, while the two NVFP4 rows share the
evaluation precision. The NVFP4+TIS
row is the standard TE NVFP4 baseline. TIS is a training correction,
not an additional inference-time operation. AIME 2024 and AIME 2025
scores average six evaluation rounds each, while MATH-500, GSM8K,
and IFEval average four rounds each. Repeated evaluations of a
checkpoint do not constitute independent training seeds. IFEval uses
Prompt-level Strict accuracy averaged across all four rounds, rather
than the best round. The evaluation records contain aggregate scores
without per-round dispersion estimates; we report point estimates
without confidence intervals.

For the three configuration rows, the final column is the unweighted arithmetic mean of the five reported
benchmark scores. Each task receives weight $1/5$, irrespective of its
evaluation-round count. The BF16 mean is 58.84\%; the NVFP4+TIS and
\minite{}+TIS means are 54.566\% and 56.350\%, displayed as 54.57\%
and 56.35\%. Parentheses report each mean minus the BF16 mean in
percentage points: $0.00$, $-4.27$, and $-2.49$. The mean gain from
naive NVFP4+TIS to \minite{}+TIS is 1.78 percentage points.
The gain row directly subtracts the NVFP4 row from the \minite{} row:
$+10.00$, $+3.88$, $+0.55$, $+0.08$, and $-5.59$ points for the five
tasks. Its final entry, $+1.78$, is the difference between their means,
not a new absolute score or a difference from BF16.

\paragraph{Task-dependent ranking.}
The IFEval ranking differs from the mathematics ranking.
The aggregate scores do not identify the cause of this difference.

\subsection{Strict-alignment diagnostic}

Strict alignment replaces learner operators with execution paths that
match the rollout implementation in formulas, kernel choices, and
reduction order. The strict configuration omits TIS, whereas the
non-strict references use it. It combines batch-invariant kernels with changes to
normalization, expert computation, and communication. This comparison
studies the resulting execution configuration rather than isolating
one deterministic-mode switch. The strict trajectory and the grouped
learner throughput measurements have separate experimental identities.

The main-text strict figure uses the exported log-probability and reward
trajectories over steps 0--1199, showing faint raw traces and centered
21-step means without exported min--max bands.
The unaligned, BF16, and aligned curves have recorded endpoints at
871, 1199, and 1189 within that window. Figure~\ref{fig:appendix-strict-full}
retains the technical report's complementary entropy and PG-loss panels,
whose original plotting protocol uses 15-step moving averages.

\IfFileExists{figures/strict_additional_metrics.pdf}{%
\begin{figure}[!htbp]
\centering
\includegraphics[width=\linewidth]{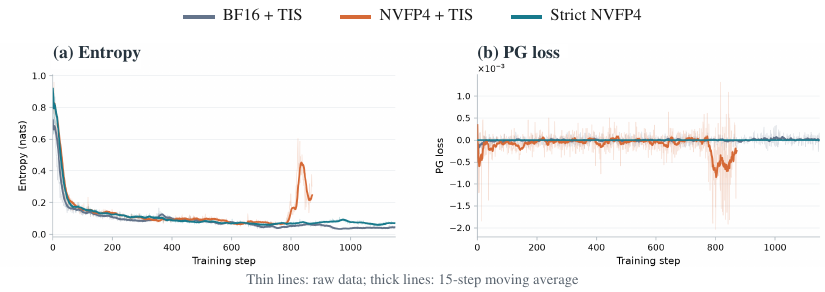}
\caption{\textbf{Additional strict-alignment training metrics.}
(a) Entropy. (b) PG loss. These two panels retain the technical
report's raw trajectories and 15-step moving averages; reward and logp
are shown in the main text. The execution configurations differ in
multiple operator choices, so this remains a joint intervention.}
\label{fig:appendix-strict-full}
\end{figure}
}{}


\section{Extended performance results}
\label{app:results-extended}

\subsection{Complete learner updates and backward precision}

\begin{table}[!htbp]
\centering
\caption{All ten independent process-trial throughputs (tokens/s).
Each entry uses the median update time over updates 6--15. The
trimmed row removes one minimum and one maximum per column; paired
speedup trims the ten within-trial ratios to BF16.}
\label{tab:appendix-trials}
\small
\setlength{\tabcolsep}{5pt}
\begin{tabular}{@{}lrrrrr@{}}
\toprule
Trial & BF16 & NVFP4 & \shortstack{CUTLASS NVFP4\\MXFP8 backward} & \shortstack{CUTLASS NVFP4\\BF16 backward} & \shortstack{\minite{}\\BF16 backward} \\
\midrule
0 & 25545 & 25863 & 30070 & 31197 & 33464 \\
1 & 24531 & 23653 & 27773 & 34662 & 31417 \\
2 & 23385 & 23291 & 27502 & 33289 & 34429 \\
3 & 23803 & 20998 & 24866 & 29219 & 28959 \\
4 & 23437 & 23473 & 26552 & 34777 & 32164 \\
5 & 25293 & 21554 & 27759 & 31109 & 32829 \\
6 & 24352 & 22382 & 26508 & 34215 & 30948 \\
7 & 22730 & 23138 & 25926 & 32079 & 33315 \\
8 & 23192 & 26673 & 28769 & 33795 & 32160 \\
9 & 26278 & 22994 & 34352 & 36279 & 33836 \\
\midrule
Trimmed mean & 24192 & 23294 & 27607 & 33140 & 32517 \\
Median & 24078 & 23215 & 27631 & 33542 & 32497 \\
Paired speedup & 1.000 & 0.959 & 1.148 & 1.369 & 1.334 \\
\bottomrule
\end{tabular}
\end{table}

The repeated experiment establishes a learner gain for both
BF16-backward variants. CUTLASS NVFP4 has slightly greater throughput than
\minite{} in this setting: 33,140 versus 32,517 tokens/s. The first three
trial-ID-paired CUTLASS NVFP4 runs give a 1.21$\times$ median gain when replacing
MXFP8 backward with BF16 backward. \minite{} with MXFP8 backward has one
corresponding trial at 27,493 tokens/s, compared with 33,464 tokens/s
for BF16 backward in trial 0, a 1.22$\times$ ratio.
The latter is a single-trial comparison.

\begin{table}[!htbp]
\centering
\caption{Optimized complete-update configuration. Each variant has one
process trial, with a trimmed mean over 13 of the 15 measured updates.
Speedups use the same-round BF16 reference. Reserved memory is the
rank-0 reported value, not a maximum over devices.}
\label{tab:appendix-optimized}
\small
\begin{tabular}{@{}lrrr@{}}
\toprule
Setting & Tokens/s & Speedup & Reserved memory (GiB) \\
\midrule
BF16 & 24134 & 1.000$\times$ & 91.61 \\
NVFP4 & 22801 & 0.945$\times$ & 99.75 \\
CUTLASS NVFP4, MXFP8 backward & 25505 & 1.057$\times$ & 107.98 \\
CUTLASS NVFP4, BF16 backward & 33065 & 1.370$\times$ & 105.54 \\
\minite{}, MXFP8 backward & 26516 & 1.099$\times$ & 106.27 \\
\minite{}, BF16 backward & \textbf{33540} & \textbf{1.390}$\times$ & 101.48 \\
\bottomrule
\end{tabular}
\end{table}

The optimized \minite{} configuration reaches 33,540 tokens/s, or
1.390$\times$ BF16 and 1.471$\times$ NVFP4. Its reserved memory is
10.8\% above BF16 and below either CUTLASS NVFP4 variant. The gain over the
repeated \minite{} setting is 3.1\%. This batch already combines the
dense path, FC1/FC2-specific GEMM-kind selection, permutation fusion,
and amax fusion in revision \texttt{b641179}; the comparison measures
their combined configuration. A separate configuration that adds
GEMM-kind selection to the older list path produced
32,264/31,174/31,896 tokens/s for BF16 backward and
24,527/24,178/23,720 tokens/s for MXFP8 backward across three trials.
Both medians are lower than the dense-path configuration. Because
the dispatch and GEMM input paths also differ, these observations
do not isolate the performance effect of GEMM-kind selection.

\subsection{Module timings and upstream implementation comparison}

The module benchmark uses one GPU and 32 local experts with balanced
synthetic dispatch. Labels 4k/16k/64k/128k denote the per-rank sequence
length before dispatch, with $1\mathrm{k}=1024$. The dispatched row
count is twice this length, giving 256/1,024/4,096/8,192 rows per
expert. Weights are warm and static. Each of five independent process
trials records 30 timing blocks of at least 10\,ms, and the table
reports the median of the five within-trial block medians. Forward
plus backward is measured directly, rather than obtained by summing
separate timing runs.
Inputs are pre-permuted, so routing and dispatch are excluded; FC1
timing includes the quantization-padding preparation used by each
quantized path, while FC2 timing starts from a precomputed activation
and excludes its padding, FC1, and weighted-SwiGLU producer.
Rows per expert are already multiples of 128 in these balanced cases.
The reported forward speedup is the geometric mean of eight separate
FC1/FC2 latency ratios, rather than the speedup of a two-layer MoE block.

\begin{table}[!htbp]
\centering
\caption{Grouped-linear timings in milliseconds across the four workload
sizes. CUTLASS NVFP4 and \minite{} use MXFP8 backward except in rows marked BF16 bwd.
TE U3045 uses its own per-token NVFP4 backward; its F+B result therefore
compares complete implementations with different backward recipes.}
\label{tab:appendix-module}
\small
\setlength{\tabcolsep}{3.7pt}
\begin{tabular}{@{}llrrrrrrrr@{}}
\toprule
& & \multicolumn{4}{c}{Forward} & \multicolumn{4}{c}{Forward + backward} \\
\cmidrule(lr){3-6}\cmidrule(l){7-10}
Layer & Setting & 4k & 16k & 64k & 128k & 4k & 16k & 64k & 128k \\
\midrule
FC1 & BF16 & .504 & .515 & .801 & 1.329 & 2.461 & 2.543 & 2.572 & 4.309 \\
& NVFP4 & 1.295 & 1.349 & 1.250 & 1.327 & 4.174 & 3.986 & 4.011 & 4.583 \\
& CUTLASS NVFP4 & .336 & .396 & .895 & 1.508 & 4.006 & 3.987 & 3.923 & 5.631 \\
& \minite{} & .372 & .424 & .582 & .932 & 3.712 & 3.830 & 3.771 & 4.989 \\
& TE U3045 & 1.128 & 1.094 & 1.275 & 1.479 & 3.873 & 3.916 & 3.846 & 5.066 \\
& \shortstack[l]{CUTLASS NVFP4\\(BF16 bwd.)} & .336 & .396 & .895 & 1.508 & 2.975 & 2.966 & 3.253 & 5.473 \\
& \shortstack[l]{\minite{}\\(BF16 bwd.)} & .372 & .424 & .582 & .932 & 2.960 & 2.990 & 3.020 & 4.870 \\
\midrule
FC2 & BF16 & .542 & .508 & .546 & .820 & 2.959 & 2.994 & 2.933 & 4.592 \\
& NVFP4 & 1.428 & 1.190 & 1.281 & 1.244 & 4.604 & 4.790 & 4.802 & 5.414 \\
& CUTLASS NVFP4 & .339 & .355 & .677 & 1.111 & 4.147 & 3.916 & 3.881 & 5.417 \\
& \minite{} & .389 & .415 & .565 & .753 & 3.875 & 3.915 & 3.907 & 5.056 \\
& TE U3045 & 1.193 & 1.043 & 1.555 & 1.747 & 4.603 & 4.507 & 4.488 & 6.234 \\
& \shortstack[l]{CUTLASS NVFP4\\(BF16 bwd.)} & .339 & .355 & .677 & 1.111 & 2.908 & 2.913 & 2.958 & 4.936 \\
& \shortstack[l]{\minite{}\\(BF16 bwd.)} & .389 & .415 & .565 & .753 & 3.160 & 2.940 & 3.090 & 4.640 \\
\bottomrule
\end{tabular}
\end{table}

\minite{} accelerates forward execution relative to TE U3045 by a geometric
mean of 2.46$\times$ across the eight cells, ranging from
1.587$\times$ to 3.065$\times$. Its F+B geometric mean is
1.10$\times$ (range 1.015--1.233$\times$), where backward recipes
differ. Relative to BF16, \minite{} forward is faster in seven of eight
cells; FC2 at 64k is approximately 0.97$\times$. \minite{} with MXFP8
backward is slower than NVFP4 at FC1 128k (0.92$\times$) and faster in
the other cells. These shape-dependent results explain why module
forward gains and complete-update gains require separate measurement.

The TE U3045 full-model runs achieve 23,124.7/23,150.0/22,345.6 tokens/s.
Their gradient norms are $10^8$--$10^{11}$, with loss still
13.7--14.5 at update 15. Since this run also quantizes nonexpert
layers and uses a different backward recipe, it serves as an
implementation reference rather than a quality-matched replacement
for the expert-only NVFP4 baseline. The benchmark loss reductions for
BF16, CUTLASS NVFP4, and \minite{} are short-window optimization
observations, separate from the RL training trajectories.

\subsection{Colocated RL and serving measurements}

The short-window RL experiment colocates slime, Megatron, and SGLang
on one eight-B300 node, using one GPU per rollout engine. The BF16
arm loads the BF16 Qwen3-30B-A3B base model and uses SGLang's Triton
MoE backend. The \minite{} arm loads its NVFP4 model, uses the dense \minite{}
expert path with BF16 backward, and combines
\texttt{flashinfer\_trtllm}, \texttt{modelopt\_fp4}, and per-token
rollout scaling. The configured rollout count is 25; checkpoint and
evaluation intervals are enlarged during timing.

\begin{table}[!htbp]
\centering
\caption{Observed components of a colocated RL iteration, in seconds.
The first two timing records are discarded. Values are arithmetic
means over BF16 steps 2--12 and \minite{} steps 2--11. The last column is
the sum of the two reported components, not the full RL iteration.}
\label{tab:appendix-rl-components}
\small
\begin{tabular}{@{}lrrrr@{}}
\toprule
Setting & Measured steps & Learner & Weight update & Sum \\
\midrule
BF16 & 11 & 47.04 & 9.94 & 56.98 \\
\minite{}, BF16 backward & 10 & 42.56 & 55.26 & 97.82 \\
\bottomrule
\end{tabular}
\end{table}

The learner timer includes log-probability computation and actor
training. The weight-update timer measures Megatron-to-SGLang
synchronization and includes BF16-to-NVFP4 conversion in \minite{}.
The learner improves by 1.11$\times$, while weight synchronization
takes 5.56$\times$ as long. Together these components cost 40.84\,s
more for these two stages of a \minite{} iteration. Their sum excludes
rollout and other iteration stages, so it establishes neither an increase
nor a reduction in total RL iteration time. Standalone serving
measurements below use a separate fixed-generation protocol.

For standalone serving, both variants use a static memory fraction of
0.6, with piecewise CUDA graphs disabled. CUDA graph batch sizes are
$1,2,4,8,16,24,32,40,48,56,64,72,80,88,96$. Real DAPO prompts are
used without input padding, with the chat template applied, a fixed
1,024-token output, and EOS termination ignored. Requests arrive at
unlimited rate with cache flushing and an 8,192-token context limit.
A single-engine sweep evaluates concurrency
$1,2,4,8,16,32,48,64,96$ with 64 prompts. Both variants select
concurrency 64; the final measurement runs all eight engines with
512 prompts.

\begin{table}[!htbp]
\centering
\caption{Eight-engine serving at concurrency 64. Throughput counts
generated output tokens, whereas learner throughput counts the fixed
tokens processed per update. These rates measure different stages.}
\label{tab:appendix-serving}
\small
\begin{tabular}{@{}lrrrr@{}}
\toprule
Setting & Output tokens/s & Requests/s & Mean TPOT (ms) & Speedup \\
\midrule
BF16 & 49119.1 & 47.97 & 10.21 & 1.000$\times$ \\
NVFP4 & 80675.8 & 78.78 & 6.05 & 1.642$\times$ \\
\bottomrule
\end{tabular}
\end{table}

Serving attains a 1.642$\times$ aggregate output-throughput gain,
and the time-per-output-token ratio is 1.69$\times$. This measurement
isolates serving capacity under fixed-length generation. The colocated
RL measurement additionally exposes weight-conversion and synchronization
costs that standalone serving omits.

\section{GB200 workload sensitivity}
\label{app:workload-sensitivity}

\paragraph{Checkpoint and input effects on expert occupancy.}
A $2\times2$ comparison crosses random initialization or pretrained
Qwen3-30B-A3B weights with mock tokens or real DAPO-Math-17k text.
All four conditions use BF16 on eight GB200 GPUs across two nodes
in one NVLink domain, with TP1/PP1/CP1/EP8, sequence length 4,096,
microbatch size 1, global batch size 64, 128 experts, and top-8 routing.
Each condition contributes one routing snapshot at step 2,
microbatch 0. Counts are aggregated over 48 layers and 128 experts,
giving 6,144 expert-layer cells per condition. An empty cell denotes
an expert that receives no token at that layer in this snapshot.

\begin{table}[!htbp]
\centering
\caption{Expert occupancy in the matched BF16 routing snapshots.
The coefficient of variation (CV) describes token counts across
experts and is averaged over layers. Counts summarize one snapshot
per condition, rather than independent training trials.}
\label{tab:appendix-workload-occupancy}
\small
\setlength{\tabcolsep}{5pt}
\begin{tabular}{@{}llrrrr@{}}
\toprule
Weights & Input & Empty cells & Empty/layer & Active/layer & Mean CV \\
\midrule
Random & Mock & 3706 (60.3\%) & 77.2 & 50.8 & 3.22 \\
Random & Real text & 4779 (77.8\%) & 99.6 & 28.4 & 3.51 \\
Pretrained & Mock & 1954 (31.8\%) & 40.7 & 87.3 & 2.78 \\
Pretrained & Real text & 89 (1.4\%) & 1.85 & 126.2 & 1.48 \\
\bottomrule
\end{tabular}
\end{table}

Real text increases the empty-cell count under random initialization
but decreases it under pretrained weights. This interaction identifies
workload variables omitted by balanced module measurements; the
snapshots do not partition latency across execution components.

\paragraph{Independent learner timing on GB200.}
The timing comparison uses the pretrained checkpoint and a 59.7-million-token
DAPO-Math-17k corpus with the model tokenizer, under the same parallelism
and batch sizes. Software revisions are \minite{} \texttt{e234f72},
TE \texttt{b07342f9}, and Megatron-LM \texttt{5215393}.
\minite{} uses native expert forward and BF16 backward. Each value
summarizes 37 updates in one process after three warm-up updates.
The initial loss of about 18.3 remains unexplained; this short window
provides timing evidence rather than policy-quality evidence.

\begin{table}[!htbp]
\centering
\caption{Complete learner-update timing on eight GB200 GPUs.
Speedup is the ratio of median update times within this comparison.}
\label{tab:appendix-gb200-timing}
\small
\begin{tabular}{@{}lrr@{}}
\toprule
Setting & Median update (ms) & Speedup \\
\midrule
BF16 & 8608.9 & 1.000$\times$ \\
\minite{}, BF16 backward & 6703.4 & 1.284$\times$ \\
\bottomrule
\end{tabular}
\end{table}

\end{document}